\pdfoutput=1

\documentclass[11pt]{article}

\usepackage[]{naacl2021}

\usepackage{graphicx}
\usepackage{times}
\usepackage{latexsym}

\usepackage[T1]{fontenc}

\usepackage[utf8]{inputenc}

\usepackage{microtype}

\title{Neural Text Classification and Stacked Heterogeneous Embeddings for Named Entity Recognition in SMM4H 2021}

\newcommand*{\affaddr}[1]{#1} 
\newcommand*{\affmark}[1][*]{\textsuperscript{#1}}

\author{Usama Yaseen\affmark[1,2], Stefan Langer\affmark[1,2]\\ 
 \affaddr{\affmark[1]Technology, Siemens AG  Munich, Germany}\\
  \affaddr{\affmark[2]CIS, University of Munich (LMU) Munich, Germany} \\
  {\tt \{usama.yaseen,langer.stefan\}@siemens.com}
}

\begin{document}
\maketitle
\begin{abstract}
This paper presents our findings from participating in the SMM4H Shared Task 2021. We addressed Named Entity Recognition (NER) and Text Classification. To address NER we explored BiLSTM-CRF with Stacked Heterogeneous Embeddings and linguistic features. We investigated various machine learning algorithms (logistic regression, Support Vector Machine (SVM) and Neural Networks) to address text classification. Our proposed approaches can be generalized to different languages and we have shown its effectiveness for English and Spanish. Our text classification submissions (team:MIC-NLP) have achieved competitive performance with F1-score of $0.46$ and $0.90$ on ADE Classification (Task 1a) and Profession Classification (Task 7a) respectively. In the case of NER, our submissions scored F1-score of $0.50$ and $0.82$ on ADE Span Detection (Task 1b) and Profession Span detection (Task 7b) respectively.
\end{abstract}

\section{Introduction}

The ubiquity of social media has led to massive user-generated content across various platforms. Twitter is a popular micro-blogging platform that allows its users to publish tweets up to 280 characters. The common public uses Twitter to share life-related personal and professional experiences with others. Personal experiences often involve health-related incidents including mentions of adverse drug effect (ADE); this information is crucial to study Pharmacovigilance. In the context of the COVID-19 pandemic, the professional experiences may include information about professions and occupations which are vulnerable due to either direct exposure to the virus or due to the associated mental health issues; detecting vulnerable occupations is critical to adopt necessary preventive measures.

Recent research focuses on mining Twitter data for adverse drug effect detection \cite{Jiang2013MiningTD, Adrover2015IdentifyingAE, onishi-etal-2018-dealing}. The distinctive style of communication on Twitter presents unique challenges including informal (brief) text, misspellings, noisy text, abbreviations, data sparsity, colloquial expressions and multilinguality.

\section{Task Description and Contribution}

\begin{figure*}[t]
    \centering
    \includegraphics[scale=0.76]{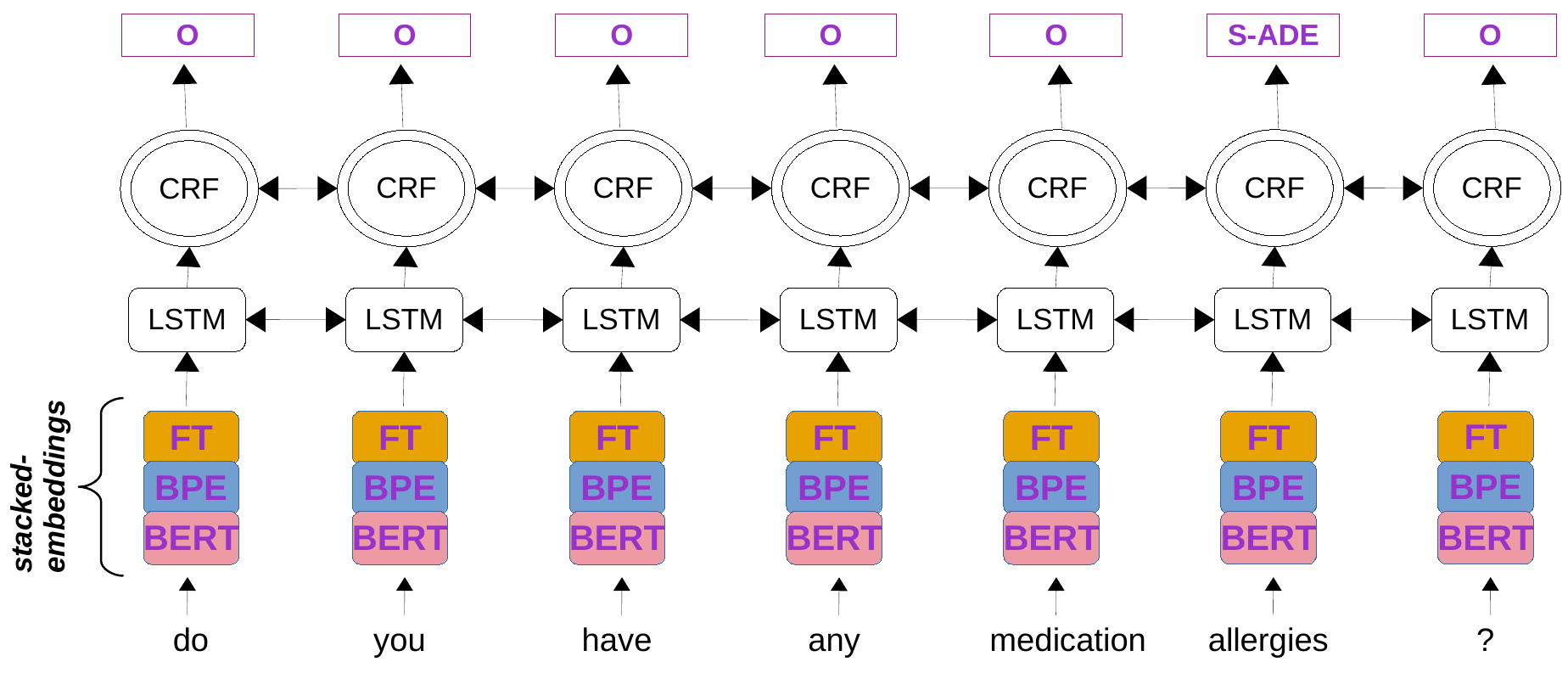}
    \caption{System architecture for NER task, consisting of BiLSTM-CRF with stacked heterogeneous embeddings. Here,  \emph{FT}: fastText embedding vector; \emph{BPE}: Byte-Pair embedding vector;  \emph{BERT}: BERT embedding vector; \emph{S\_ADE}: S\_Adverse Drug Effect.}
    \label{fig:model-ner}
\end{figure*}

We participate in the following two tasks organized by SMM4H workshop 2021 \cite{magge2021overview}: (1) Task 1: Classification, Extraction and Normalization of Adverse Effect mentions in English tweets (2) Task 7: Identification of professions and occupations in Spanish tweets \cite{miranda2021profner}. Task 1 consists of three sub-tasks, (a): ADE tweet classification, (b): ADE span detection, (c): ADE resolution; whereas Task 7 consists of two sub-tasks: (a): Tweet classification (b): Profession/occupation span detection. For both tasks, we participate in sub-tasks (a) and (b). The Task 1a and Task 7a is a text classification problem while Task 1b and Task 7b is a Named Entity Recognition problem.

Following are our multi-fold contributions:

1. To address NER tasks, we have employed a neural network based sequence classifier, i.e. BiLSTM-CRF and investigated various heterogeneous embeddings. We further investigated the combination of character embeddings, static word embeddings and contextualized embeddings in a stacked format. We also incorporated linguistic features such as part-of-speech tags (POS), orthographic features etc. We apply the proposed modelling approaches to both English and Spanish texts. In {\it Profession span detection} (Task 7b) our submission (team:MIC-NLP) achieved the F1-score of $0.824$ which is $6$ points higher than the arithmetic median of all the submissions; in case of {\it ADE span detection} our submission scored F1-score of $0.50$, around $8$ points higher than the arithmetic median of the participating submissions.

2. To address text classification tasks, we investigated various machine learning algorithms like {\it logistic regression}, {\it SVM} and {\it neural network} with various word and sentence embeddings. In {\it ADE tweet classification} (Task 1a) our submission (team:MIC-NLP) scored F1-score of $0.46$, approximately $2$ points higher than the arithmetic median of participating submissions; in case of {\it tweet classification} (task 7a) our system achieved the F1-score of $0.90$ which is $5$ points higher than the arithmetic median of all submissions. 

\section{Methodology}

In the following sections we discuss our proposed model for named entity recognition and text classification.

\subsection{Named Entity Recognition}

Figure \ref{fig:model-ner} describes the architecture of our model, where we design a sequence tagger to extract entities. The architecture of our model is a standard BiLSTM-CRF \cite{lample-etal-2016-neural} model with stacked heterogeneous embeddings and linguistic features as input. The stacked embeddings consists of Byte-Pair subword embeddings \cite{heinzerling2018bpemb}, fastText subword embeddings \cite{BojanowskiGJM17} and contextualized word embeddings \cite{devlin-etal-2019-bert, liu-2019-roberta}. The linguistic features include POS, capitalization features and orthographic features.

\subsection{Text Classification}

We explored traditional machine learning algorithms like logistic regression, SVM and neural network based architecture with various word and sentence embeddings for text classification. The SVM was trained with Radial Basis Function (RBF) Kernel with the value of penalty parameter C determined by grid search for each dataset. Our best model was a Neural Network with contextualized embeddings \cite{devlin-etal-2019-bert, liu-2019-roberta}. Since both datasets (Task 1a and Task 7a) were highly imbalanced, we employed higher class weights for minority classes to train the final models.

\begin{table}[t]
\center
\renewcommand*{\arraystretch}{1.2}
\resizebox{.36\textwidth}{!}{
\begin{tabular}{l | c | c}
\textbf{Task} & \textbf{Train} & \textbf{Dev} \\ \hline
\multicolumn{3}{c}{\textbf{Sentence Counts}}  \\ \hline
Task 1b & 34142 & 1775 \\ \hline
Task 7b  & 14755 & 4959 \\ \hline
\multicolumn{3}{c}{\textbf{Task 1b Entities}}  \\ \hline
ADE & 1713 & 87 \\ \hline
\multicolumn{3}{c}{\textbf{Task 7b Entities}}  \\ \hline
PROFESION & 1597 & 566 \\ \hline
SITUACION\_LABORAL & 264 & 85 \\ \hline
ACTIVIDAD & 45 & 16 \\ \hline
FIGURATIVA & 16 & 8
\end{tabular}
}
\caption{Dataset statistics for NER.}
\label{table:ner-statistics}
\end{table}

\begin{table}[t]
\centering
\resizebox{.36\textwidth}{!}{
\begin{tabular}{l | c}
\hline
\textbf{Hyper-parameter} & \textbf{Value} \\ \hline
\multicolumn{2}{c}{{\bf NER}} \\ \hline
learning rate & $0.1$ \\ \hline
optimizer & SGD \\ \hline
hidden size & $256$ \\ \hline
POS dimensions &$50$ \\ \hline
Ortho dimension &$50$ \\ \hline
batch size & $32$ \\ \hline
epochs & $150$ \\ \hline
\multicolumn{2}{c}{{\bf Text Classification}} \\ \hline
kernel & RBF \\ \hline
class-weights & $10.0$ \\ \hline
learning rate & $0.00003$ \\ \hline
batch size & $16$ \\ \hline
epochs & $10$ \\ \hline
\end{tabular}
}
\caption{Hyper parameter settings for NER and Text classification.}
\label{table:hyper-params}
\end{table}

\subsection{Ensemble Strategy}

Bagging is a useful technique to reduce the variance of the learning algorithm without impacting bias.
We employed a variant of Bagging \cite{Br:96}  such that every data point in the training set is part of the development set at least once and vice versa. We created three data folds and trained the model using optimal configuration on each fold, inference on the test set involves majority voting among the three trained models.

For NER, we perform majority voting at the token level for each test data point. In cases when voting results in a tie, we take the prediction of the confident model, we treat the model trained on original data split as the
confident model. In the case of an ensemble for text classification, we followed the straight forward approach of majority voting at sentence level for each test data point.

\label{section:bagging}

\section{Experiments and Results}

\subsection{Dataset and Experimental Setup}

{\bf Data:} We employed bagging (discussed in section \ref{section:bagging}) to split the annotated corpus into 3-folds. For ADE span detection (Task 1b) and Profession span detection (Task 7b) we perform sentence splitting, word tokenization, computing orthographic features and POS tagging. We do not perform any pre-processing for ADE classification (Task 1a) and Tweet classification (Task 7a).

{\it ADE Classification (Task 1a):} The dataset consists of tweets in the English language and the task is to detect tweets containing adverse drug effect. The dataset contains two classes, {\it ADE} and {\it NoADE}. The dataset is highly imbalanced with only $1235$ tweets of type ADE out of total $17385$ tweets in the train set.

{\it ADE Span Detection (Task 1b):} The dataset consists of only one entity type {\it ADE}. The train set contains $1717$ entity mentions of {\it ADE} (see Table \ref{table:ner-statistics}).

{\it Profession Classification (Task 7a):} The dataset consists of tweets in the Spanish language and the task is to detect tweets containing mention of profession/occupation. The dataset contains two classes. The dataset is highly imbalanced with only $1393$ tweets containing a positive mention out of $6000$ tweets.

{\it Profession Span Detection (Task 7b):} The dataset consists of four entity types with few mentions of type {\it FIGURATIVA} as shown in Table \ref{table:ner-statistics}.  Entities of type ACTIVIDAD and FIGURATIVA are ignored in the evaluation of this shared task but we still treat them as regular entities.

{\bf Experimental Setup:} We found contextualized embeddings to be very helpful in identifying entities and text classification; all our experiments used pre-trained contextualized embeddings. We employ {\it RoBERTa} \cite{GururanganMSLBD20} for Task 1a and Task 1b; we use multi-lingual BERT \cite{devlin-etal-2019-bert} for Task 7a and Spanish BERT \cite{CaneteCFP2020} for Task 7b. We do not finetune embeddings in our experiments. We don’t employ any strategy for handling imbalanced classes for NER but have used class weighting by a factor of 10 for all positive classes for text classification. Table \ref{table:hyper-params} lists the best configuration of hyperparameters for all the tasks.

\subsection{Results on Development Set}

\begin{table}[t]
\centering
\resizebox{.45\textwidth}{!}{
\begin{tabular}{r | l | c | c}
& \textbf{Features} & \textbf{Task 1b} & \textbf{Task 7b} \\
& & P/R/F1 & P/R/F1 \\ \hline
r1 & {\it glove}  & $.5/.18/.26$ & - \\ 
r2 &  {\it fastText} & $.89/.28/.43$ & $.84/.64/.73$ \\ 
r3 &  {\it fastText + Char} & $.64/.28/.39$ & $.83/.67/.74$\\ 
r4 &  {\it fastText + BytePair} & $.62/.34/.44$ & $.82/.74/.78$\\ 
r5 &  {\it BERT} & $.68/.35/.46$ & $.84/.76/.80$\\ 
r6 &  {\it BERT + fastText + BytePair} & ${\bf .61/.52/.56}$ & ${\bf .86/.77/.81}$ \\ 
 &   & {\bf Fold=2} & {\bf Fold=2} \\ 
r7 &  {\it BERT + fastText + BytePair} & ${\bf .80/.21/.34}$ & ${\bf .85/.79/.82}$ \\
 &   & {\bf Fold=3} & {\bf Fold=3} \\ 
r8 &  {\it BERT + fastText + BytePair} & ${\bf .77/.37/.50}$ & ${\bf .84/.78/.81}$ \\

\end{tabular}
}
\caption{Scores on dev set using different features for {\it BiLSTM-CRF} on {\it Task 1b} and {\it Task 7b}.}
\label{table:ner}
\end{table}

\begin{table}[t]
\centering
\resizebox{.45\textwidth}{!}{
\begin{tabular}{r | l | c | c}
& \textbf{Features} & \textbf{Task 1a} & \textbf{Task 7a} \\
& & P/R/F1 & P/R/F1 \\ \hline
r1 & {\it logisticReg + fastTextSentEmb}  & $.33/.83/.47$ & $.38/.95/.55$ \\ 
r2 &  {\it logisticReg + BERTSentEmb} & $.34/.81/.48$ & $.41/.83/.55$\\ 
r3 &  {\it logisticReg + BERTWordEmbSum} & $.45/.86/.59$ & $.45/.86/.59$\\ 
r4 & {\it SVM + fastTextSentEmb}  & $.53/.66/.59$ & $.71/.67/.69$ \\ 
r5 &  {\it SVM + BERTSentEmb} & $.36/.86/.51$ & $.49/.66/.56$\\ 
r6 &  {\it SVM + BERTWordEmbSum} & $.44/.90/.59$ & $.61/.64/.63$\\
r7 &  {\it NeuralNetwork + Glove} & $.51/.63/.56$ & $.64/.59/.61$\\
r8 &  {\it NeuralNetwork + BERT} & ${\bf .77/.72/.74}$ & ${\bf .95/.85/.90}$ \\
r9 &  & {\bf Fold=2} & {\bf Fold=2} \\
r10 &  {\it NeuralNetwork + BERT} & ${\bf .79/.66/.72}$ & ${\bf .89/.91/.90}$ \\
r11 &  & {\bf Fold=3} & {\bf Fold=3} \\
r12 &  {\it NeuralNetwork + BERT} & ${\bf 0.8/.65/.72}$ & ${\bf .93/.84/.88}$ \\

\end{tabular}
}
\caption{Scores on dev set using different features on {\it Task 1a} and {\it Task 7a}.}
\label{table:clf}
\end{table}

We perform various experiments to investigate the impact of features on performance on the development set.

{\bf NER:} Table \ref{table:ner} shows the score on the development set for Task 1b and Task 7b. Observe that fastText embeddings (row r2) outperform glove embeddings (row r1) for Task 1b. Subsequently, fastText embeddings with BytePair embeddings (row r4) provide an improvement over only fastText (row r2) and the combination of fastText with Character embeddings (row r3). The contextualized embeddings (row r5) provide an improvement over the combination of fastText with BytePair embeddings. In row r6, we employ BERT, fastText and BytePair embeddings in a stacked format leading to the best f1-score for both Task 1b and Task 7b.

{\bf Text Classification:} Table \ref{table:clf} shows the score on the development set for Task 1a and Task 7a. Observe that BERTSentEmb provides improvement over fastTextSentEmb for both logistic regression and SVM. Similarly, BERTWordEmbSum further improves BERTSentEmb. BERTSentEmb uses BERT's {\it CLS} representation whereas BERTWordEmbSum is computed by average of the token-wise embeddings of pre-trained BERT as discussed in \citeauthor{RogersKR20}. Neural Network with BERT achieves the best result for both datasets.

\subsection{Results on Test Set}

Table \ref{table:test-results} shows the comparison of our submissions with the arithmetic median of the participating teams for all the tasks. Our submissions achieve the overall best F1-score than the arithmetic median for all the tasks showing compelling advantage. For Task 1a, the precision of our system is lower than the arithmetic median but this is compensated by the improvement in recall. For all the tasks, the precision is higher than the recall but overall precision and recall are balanced.

\begin{table}[t]
\centering
\resizebox{.45\textwidth}{!}{
\begin{tabular}{r | l | c | c}
& \textbf{Tasks} & {\bf Arithmetic Median} & \textbf{MIC-NLP} \\
& & {\it P/R/F1} & {\it P/R/F1} \\ \hline
r1 & {\it Task 1a}  & ${\bf .50}/.40/.44$ & $.47/{\bf .45/.46}$ \\ 
r2 & {\it Task 1b}  & $.49/.45/.42$ & ${\bf .55/.45/.50}$ \\ 
r3 & {\it Task 7a}  & $.91/.85/.85$ & ${\bf .94/.85/.90}$ \\ 
r4 & {\it Task 7b}  & $.84/.72/.76$ & ${\bf .85/.79/.82}$ \\ 

\end{tabular}
}
\caption{Comparison of our system (team:MIC-NLP) with the arithmetic median of the participating teams. Scores on test set for Task 1a, Task 1b, Task 7a and Task 7b.}
\label{table:test-results}
\end{table}

\section{Conclusion}

In this paper, we described our system with which we participate in Task 1(Adverse Drug Effect Classification and Extraction) and Task 7 (Identification of professions and occupations in Spanish Tweets) in the SMM4H Shared Task 2021. Our NER system employed stacked heterogeneous embeddings to extract entities in English and Spanish text. Our NER system demonstrates a competitive performance with F1-score of $0.50$ and $0.82$ on ADE Span Detection (Task 1b) and Profession/Occupation span detection (Task 7b) respectively. Our text classification system employed contextualized embeddings with Neural Network as a classifier to achieve a competitive performance with F1-score of $0.46$ and $0.90$ on ADE Classification (Task 1a) and Profession/Occupation classification (Task 7a) respectively. In future, we would like to improve error analysis to further enhance our NER and text classification models.

\section*{Acknowledgment}
This research was supported by Bundesministerium für Wirtschaft und Energie (bmwi.de), grant 01MD19003E (PLASS,  plass.io) at Technology - Siemens AG, Munich Germany.

\bibliography{anthology,custom}
\bibliographystyle{acl_natbib}

\end{document}